\pdfoutput=1
\documentclass[conference]{IEEEtran}
\IEEEoverridecommandlockouts

\usepackage{amsmath}
\usepackage{graphicx}
\usepackage{booktabs}
\usepackage{multirow}
\usepackage{xcolor}
\usepackage{tikz}
\usetikzlibrary{positioning,arrows.meta,calc}
\usepackage{cite}
\usepackage[hidelinks]{hyperref}

\newif\ifshownotes
\shownotesfalse 
\ifshownotes
  \DeclareRobustCommand{\TODO}[1]{{\color{red}\textbf{[TODO:} #1\textbf{]}}}
  \DeclareRobustCommand{\AUTHORCHECK}[1]{{\color{blue}\textbf{[AUTHOR CHECK:} #1\textbf{]}}}
  
\else
  \DeclareRobustCommand{\TODO}[1]{\PackageWarning{draftnotes}{Unresolved TODO}}
  \DeclareRobustCommand{\AUTHORCHECK}[1]{\PackageWarning{draftnotes}{Unresolved AUTHORCHECK}}
  
\fi

\newcommand{\none}{\textsc{None}}
\newcommand{\flow}{\textsc{Flow}}
\newcommand{\geom}{\textsc{Geom}}
\newcommand{\hyp}[1]{\textbf{H#1}}
\newcommand{\NFAST}{\bar{N}_{\mathrm{F}}}
\newcommand{\Hgrad}{\bar{H}_{\nabla}}

\hypersetup{
  pdftitle={Does Dynamic-Point Filtering Help When Texture Is Scarce? A Controlled Study of ORB-SLAM2 Front-Ends in Synthetic Indoor Scenes},
  pdfauthor={Zekui Xue}
}

\begin{document}

\title{Does Dynamic-Point Filtering Help When Texture Is Scarce?\\
A Controlled Study of ORB-SLAM2 Front-Ends\\ in Synthetic Indoor Scenes}

\author{%
\IEEEauthorblockN{Zekui Xue}
\IEEEauthorblockA{Department of Computer Science\\
University of Bath\\
Bath, United Kingdom\\
\texttt{zekui.xue@bath.edu}}
}

\maketitle

\begin{abstract}
Dynamic-point filters are routinely added to feature-based visual SLAM, and several recent systems argue that removing dynamic features can leave too few static features in low-texture regions. So far, these systems have been evaluated only on texture-rich benchmark sequences. We present a controlled study that isolates this interaction. We render synthetic indoor sequences in which surface texture (four levels, quantified by FAST-corner density and image-gradient entropy) and scene dynamics (three levels) are varied factorially along identical camera trajectories, with stereo, RGB-D, ground-truth poses and dynamic masks. On this grid we compare ORB-SLAM2 without filtering, with an optical-flow and epipolar-residual filter (FLOW), and with a multi-view depth-consistency filter (GEOM), and report trajectory error, tracking completeness and surviving static features over five runs. Because the masks give per-keypoint ground truth, we also measure each filter's dynamic-point precision and recall and its static-feature false-removal rate, so that mechanistic explanations can be tested directly. We do not propose a new filter. On 720 runs over 24 sequences, filtering helped mainly in the most dynamic cells; the benefit did not decline monotonically with texture, but at the lowest level filtering reduced tracking completeness, and ORB-SLAM2 never initialised in static L3 scenes. Contrary to our hypothesis, GEOM discarded more static keypoints than FLOW (median FRR 6.8\% vs.\ 1.5\% for RGB-D, 19.6\% vs.\ 1.5\% for stereo); its RGB-D advantage tracked dynamic-point recall and vanished in stereo mode. Data and code are available at \url{https://github.com/felixxxue/texture-dynamics-slam}.
\end{abstract}

\begin{IEEEkeywords}
Visual SLAM, dynamic environments, low-texture scenes, ORB-SLAM2, synthetic data, evaluation
\end{IEEEkeywords}

\section{Introduction}
\label{sec:intro}

Feature-based visual SLAM systems such as ORB-SLAM and ORB-SLAM2~\cite{murartal2015orbslam,murartal2017orbslam2} estimate the camera pose from keypoints that are assumed to lie on static structure. Moving people and objects violate this assumption, and a large family of dynamic-SLAM front-ends therefore removes keypoints that are judged to be dynamic before, or during, tracking. Two classical cues for this decision are \emph{optical flow with an epipolar consistency test}, as in the moving-consistency check of DS-SLAM~\cite{yu2018dsslam}, and \emph{multi-view geometry}, as in the depth and parallax test of DynaSLAM~\cite{bescos2018dynaslam}; both were built on ORB-SLAM2.

Filtering has a cost that is easy to overlook in texture-rich scenes: every removed keypoint, whether truly dynamic or a false positive, is one fewer constraint for pose estimation. When the scene offers few keypoints in the first place, for example white walls, uniform floors and plain furniture, the same filter may push the tracker below the number of static features it needs. Evaluation studies have shown that sparse feature-based systems such as ORB-SLAM2 and ORB-SLAM3 are severely affected in textureless environments~\cite{bujanca2021robust}. Three recent systems state the interaction explicitly: IL-SLAM notes that dynamic removal creates the new problem of insufficient remaining point features and adds line features when needed~\cite{zhang2025ilslam}; SR-SLAM scores frame reliability partly by feature richness~\cite{zhang2025srslam}; and SFSE-SLAM observes that ``removing dynamic feature points often leaves insufficient static features in low-texture regions'' and compensates with a density-driven adaptive threshold~\cite{li2026sfse}. However, these systems are evaluated on TUM RGB-D~\cite{sturm2012tum} and Bonn~\cite{palazzolo2019refusion} sequences (and, for SR-SLAM, on real robot runs), none of which varies texture in a controlled way. Whether, and by how much, the benefit of dynamic-point filtering shrinks as texture becomes scarce has, to our knowledge, not been measured.

This paper measures it. We hold scene geometry, camera trajectory and the motion of dynamic agents fixed, vary surface texture and the number of moving agents factorially in simulation, and run ORB-SLAM2 with no filter, a flow-based filter and a geometry-based filter in both stereo and RGB-D modes. \textbf{We do not propose a new filter}; our aim is to quantify, under controlled texture, when existing filtering principles help or hurt, and why.

\textbf{Contributions.}
\begin{enumerate}
\item[C1] \emph{Data.} We release 24 synthetic indoor sequences in which surface texture (four levels, quantified by a reproducible texture metric) and scene dynamics (three levels) are varied factorially along identical camera trajectories, with stereo images, RGB-D, ground-truth poses and per-pixel dynamic masks (Sec.~\ref{sec:data}).
\item[C2] \emph{Interaction.} Using these data, we quantify how the effect of flow-based and geometry-based dynamic-point filtering on ORB-SLAM2 changes with texture, reporting ATE, RPE, tracking completeness and surviving static features over $N_r=5$ runs per configuration (Sec.~\ref{sec:grid}).
\item[C3] \emph{Mechanism.} Using the ground-truth masks, we measure each filter's per-keypoint dynamic-point precision and recall and its static-feature false-removal rate, and test whether these explain the accuracy differences between filters (Sec.~\ref{sec:mechanism}); in our data they do, but not in the way we hypothesised. As an additional analysis, we compare stereo and RGB-D input, whose depth sources depend differently on texture (Sec.~\ref{sec:sensor}).
\end{enumerate}

\section{Related Work}
\label{sec:related}

\textbf{Dynamic visual SLAM.} DynaSLAM~\cite{bescos2018dynaslam} extends ORB-SLAM2 (monocular, stereo and RGB-D) with Mask R-CNN segmentation and, for RGB-D, a multi-view geometry test that projects points from overlapping keyframes and flags them by depth discrepancy and parallax angle; it also inpaints the occluded background. DS-SLAM~\cite{yu2018dsslam} combines ORB-SLAM2 with SegNet and a moving-consistency check that tracks features with Lucas--Kanade optical flow~\cite{lucas1981iterative} and tests them against the epipolar constraint. Detect-SLAM~\cite{zhong2018detect} runs an SSD detector on keyframes and propagates moving probabilities to map points. FlowFusion~\cite{zhang2020flowfusion} segments dynamic regions from PWC-Net optical flow in dense RGB-D SLAM, and ReFusion~\cite{palazzolo2019refusion} detects dynamics from TSDF residuals and released the Bonn RGB-D Dynamic dataset. DynaPix SLAM~\cite{xu2023dynapix} estimates per-pixel motion probabilities without semantics and feeds them into ORB-SLAM2's map-point selection and bundle adjustment, evaluating on the synthetic GRADE data and on TUM RGB-D. Our \flow{} and \geom{} filters (Sec.~\ref{sec:filters}) are generic, semantics-free instances of the DS-SLAM and DynaSLAM geometric principles; we study them as isolated factors rather than as complete systems.

\textbf{Low-texture SLAM.} Feature scarcity is addressed by point-and-line systems~\cite{pumarola2017plslam,gomezojeda2019plslam}, by structural and planar constraints~\cite{li2021structural,yunus2021manhattan}, by dedicated weak-texture pipelines such as RWT-SLAM~\cite{peng2022rwtslam}, by direct and semi-direct methods~\cite{engel2018dso,fontan2023sidslam}, and by learned dense methods such as DROID-SLAM~\cite{teed2021droid}. SID-SLAM~\cite{fontan2023sidslam} also released the DLR Minimal Texture dataset of real RGB-D sequences with motion-capture ground truth. In these works texture is typically a \emph{categorical} condition (e.g., the TUM structure/texture sequences~\cite{sturm2012tum}, OpenLORIS corridors~\cite{shi2020openloris} or dedicated minimal-texture recordings) rather than a continuous, controlled variable. Evaluation studies~\cite{bujanca2021robust,sharafutdinov2021comparison} compare systems across challenge categories; Bujanca et al.~\cite{bujanca2021robust} report texture and dynamics as separate categories, not crossed.

\textbf{Feature-scarcity-aware dynamic SLAM.} IL-SLAM~\cite{zhang2025ilslam} triggers line features when a feature-awareness criterion indicates that too few points remain after dynamic removal; it is evaluated on TUM sequences. SR-SLAM~\cite{zhang2025srslam} computes a frame reliability score from detection confidence, spatial distribution, feature richness on a $3\times3$ grid and depth quality, and falls back to direct photometric and depth refinement when features are scarce; it is evaluated on TUM, Bonn and real robot runs. SFSE-SLAM~\cite{li2026sfse} adds texture-aware compensation through a threshold driven by global feature density, evaluated on TUM and Bonn. These papers motivate our question; our study is complementary, since it does not add a mechanism but measures the underlying effect under controlled texture.

\textbf{Synthetic data and benchmarks.} ICL-NUIM~\cite{handa2014iclnuim} provides static synthetic indoor scenes with largely low-texture walls. TartanAir~\cite{wang2020tartanair} provides photo-realistic sequences with stereo, depth, segmentation, optical flow and moving objects. VIODE~\cite{minoda2021viode} renders each of three environments at four dynamic levels that progressively add moving objects, with instance masks. GRADE~\cite{bonetto2023grade} generates indoor dynamic scenes with animated humans in Isaac Sim, and a companion study~\cite{bonetto2023simulation} found that DynaSLAM achieved good ATE on such data while losing track for at least about 27\,s, i.e.\ ATE alone can hide failures. The CSE dataset~\cite{park2024cse} provides static and dynamic versions of service environments with identical trajectories. Robust-SLAM~\cite{xu2024perturbation} perturbs Replica scenes with image, depth and trajectory corruptions under an explicitly static-world assumption, and VSLAM-LAB~\cite{fontan2025vslamlab} offers a unified pipeline for re-running baselines. Table~\ref{tab:positioning} summarises which of these control texture and dynamics; in the sources we consulted we found none that varies both factorially along the same trajectory.

\begin{table}[t]
\caption{Positioning. ``no'' means that no explicit, graded control is reported in the sources we consulted; n/s: not stated in those sources.}
\label{tab:positioning}
\centering
\scriptsize
\setlength{\tabcolsep}{2.5pt}
\begin{tabular}{@{}lcccc@{}}
\toprule
Dataset / study & Synth. & Texture ctrl. & Dynamics ctrl. & Dyn.\ masks \\
\midrule
TUM RGB-D~\cite{sturm2012tum}            & no    & categ.$^{a}$ & categ.$^{b}$ & no \\
Bujanca et al.~\cite{bujanca2021robust}  & mixed & separate$^{c}$ & separate$^{c}$ & n/s \\
TartanAir~\cite{wang2020tartanair}       & yes   & no & not graded & seg.\ only$^{d}$ \\
VIODE~\cite{minoda2021viode}             & yes   & no & 4 levels & yes \\
GRADE~\cite{bonetto2023grade}            & yes   & no & n/s & n/s \\
CSE~\cite{park2024cse}                   & yes   & no & static/dyn.\ twins & n/s \\
Robust-SLAM~\cite{xu2024perturbation}    & yes   & no & static only & -- \\
IL-/SR-/SFSE-SLAM$^{e}$ & no & no & no & n/s \\
\midrule
\textbf{This work}                       & yes   & 4 levels + metric & 3 levels & yes \\
\bottomrule
\end{tabular}\\[2pt]
\parbox{\columnwidth}{\scriptsize $^{a}$Static structure/texture sequences. $^{b}$Separate sitting/walking recordings. $^{c}$Reported as separate challenge categories, not crossed. $^{d}$Segmentation is provided; dynamic-point labels are not stated. $^{e}$Evaluations reported in~\cite{zhang2025ilslam,zhang2025srslam,li2026sfse} (TUM RGB-D, Bonn, real robot runs).}
\end{table}

\section{Study Design}
\label{sec:design}

\subsection{Backbone}
\label{sec:backbone}
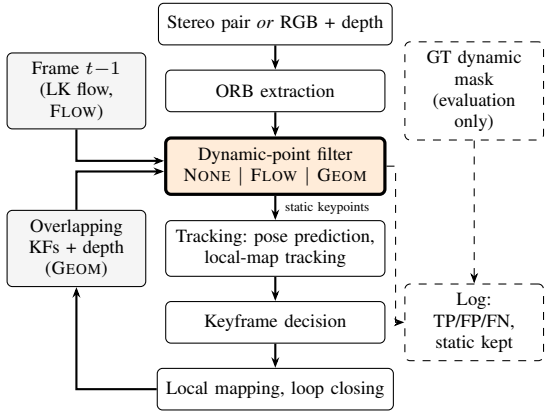
\begin{figure}[t]
\centering
\begin{tikzpicture}[
  font=\scriptsize,
  box/.style={draw, rounded corners=2pt, minimum width=2.9cm, minimum height=0.55cm, align=center, fill=white},
  side/.style={draw, rounded corners=2pt, text width=1.6cm, minimum height=0.55cm, align=center, fill=gray!8},
  filt/.style={box, fill=orange!15, very thick},
  arr/.style={-{Stealth[length=4pt]}, thick},
  darr/.style={-{Stealth[length=4pt]}, dashed},
  node distance=0.32cm]
  \node[box] (in) {Stereo pair \emph{or} RGB + depth};
  \node[box, below=of in] (orb) {ORB extraction};
  \node[filt, below=of orb] (f) {Dynamic-point filter\\ \none{} $|$ \flow{} $|$ \geom{}};
  \node[box, below=of f] (tr) {Tracking: pose prediction,\\ local-map tracking};
  \node[box, below=of tr] (kf) {Keyframe decision};
  \node[box, below=of kf] (lm) {Local mapping, loop closing};
  \node[side, left=0.25cm of orb] (prev) {Frame $t{-}1$\\(LK flow, \flow)};
  \node[side, left=0.25cm of tr] (kfs) {Overlapping KFs + depth (\geom)};
  \node[side, right=0.25cm of orb, dashed, fill=white] (gt) {GT dynamic mask\\(evaluation only)};
  \node[side, right=0.25cm of kf, dashed, fill=white] (log) {Log: TP/FP/FN,\\ static kept};
  \draw[arr] (in) -- (orb);
  \draw[arr] (orb) -- (f);
  \draw[arr] (f) -- node[right, font=\tiny]{static keypoints} (tr);
  \draw[arr] (tr) -- (kf);
  \draw[arr] (kf) -- (lm);
  \draw[arr] (prev.south) |- ([yshift=2pt]f.west);
  \draw[arr] (kfs.north) |- ([yshift=-2pt]f.west);
  \draw[arr] (lm.west) -| (kfs.south);
  \draw[darr] (gt.south) -- (log.north);
  \draw[darr] (f.east) -- ++(0.12,0) |- (log.west);
\end{tikzpicture}
\caption{Where the dynamic-point filter sits in the ORB-SLAM2 tracking front-end (schematic drawn for this paper). Flagged keypoints are removed before tracking and map-point creation. Dashed elements are used only for evaluation (Sec.~\ref{sec:metrics}) and never influence the tracker.}
\label{fig:pipeline}
\end{figure}

We use the public ORB-SLAM2 implementation~\cite{murartal2017orbslam2} in its stereo and RGB-D modes with the default tracking, local-mapping and loop-closing threads and with $N_f=1000$ ORB features~\cite{rublee2011orb} per frame (scale factor 1.2, 8 levels, FAST thresholds 20/7, close/far depth threshold 40 baselines); all other parameters are left at their defaults. We use commit \texttt{f2e6f51} of the public repository, built headless (no viewer) on Windows with MSVC~19.44 and OpenCV~4.10; the port changes only portability code (OpenCV~4 headers, a variable-length array, an allocator type, sleeping) and adds a binary vocabulary loader, and the released patch lists every change. A dynamic-point filter is inserted between ORB extraction and tracking (Fig.~\ref{fig:pipeline}): keypoints flagged as dynamic are excluded from pose prediction, local-map tracking (motion-only bundle adjustment) and map-point creation. The filter only removes keypoints; it does not add features, change the ORB extractor, or alter the back-end. For evaluation only, each keypoint decision is logged together with its ground-truth label.

\subsection{Dynamic-point filters}
\label{sec:filters}
Both filters produce a binary label $\hat{y}_k\in\{0,1\}$ (1 = dynamic) for every ORB keypoint $\mathbf{x}_k$ of the current frame $t$. Homogeneous coordinates are denoted by $\tilde{\cdot}$.

\textbf{\flow{}: optical flow with an epipolar residual test} (after the moving-consistency check of DS-SLAM~\cite{yu2018dsslam}, without semantic segmentation). Each keypoint $\mathbf{x}_k$ of frame $t$ is tracked backwards into frame $t{-}1$ with pyramidal Lucas--Kanade flow~\cite{lucas1981iterative} ($w_{\mathrm{LK}}=21$\,px window, $\ell_{\mathrm{LK}}=3$ levels), giving $\mathbf{x}'_k$. Pairs with a large LK error ($>e_{\mathrm{LK}}=20$, mean absolute patch difference) or within $b=10$\,px of the image border are discarded and left unlabelled ($\hat{y}_k=0$). A fundamental matrix $\mathbf{F}_t$ is estimated from the remaining pairs with RANSAC~\cite{hartley2004mvg} (OpenCV \texttt{findFundamentalMat}, inlier threshold $\tau_F=1$\,px, confidence 0.99); with fewer than 15 pairs the frame is not filtered. With the epipolar line $\mathbf{l}_k=\mathbf{F}_t\tilde{\mathbf{x}}'_k=(l_1,l_2,l_3)^\top$, the residual is
\begin{equation}
d_k=\frac{|\tilde{\mathbf{x}}_k^\top \mathbf{F}_t\tilde{\mathbf{x}}'_k|}{\sqrt{l_1^2+l_2^2}},\qquad \hat{y}_k=[\,d_k>\tau_{\mathrm{epi}}\,],
\label{eq:epi}
\end{equation}
with $\tau_{\mathrm{epi}}=1$\,px. Optionally, flagged keypoints are dilated into a region mask of radius $r_f$ and all keypoints inside it are removed; $r_f=0$ gives a purely point-wise filter, $r_f>0$ a region-level filter as used by mask-based methods. We report the point-wise variant, $r_f=0$. Note that $\mathbf{F}_t$ is estimated from the same keypoints it is used to test, so its quality depends on the number and spatial spread of static correspondences.

\textbf{\geom{}: multi-view depth consistency} (after the geometric branch of DynaSLAM~\cite{bescos2018dynaslam}, without Mask R-CNN). Given a preliminary pose $\hat{\mathbf{T}}_t$ from ORB-SLAM2's standard tracking with its robust kernel, we select the $K_o=5$ keyframes with the largest overlap (shared map points) with frame $t$. Every keyframe keypoint with valid depth is back-projected to a 3D point $\mathbf{X}$ and projected into frame $t$, giving pixel $\mathbf{x}'$ and projected depth $z_{\mathrm{proj}}$. Let $\alpha$ be the angle between the viewing rays of $\mathbf{X}$ from the keyframe and from frame $t$, and $z'$ the depth measured in frame $t$ at $\mathbf{x}'$; to avoid seeds caused by sub-pixel projection errors at depth edges, $z'$ is the \emph{largest} valid depth in a $5\times5$ patch around $\mathbf{x}'$ (RGB-D) or among current keypoints with stereo depth within 3\,px of $\mathbf{x}'$ (stereo). Points with $\alpha>\alpha_{\max}=30^\circ$ are skipped because of possible self-occlusion; otherwise
\begin{equation}
\Delta z=z_{\mathrm{proj}}-z',\qquad \mathbf{x}'\ \text{is a dynamic seed if}\ \Delta z>\tau_z,
\label{eq:geom}
\end{equation}
with $\tau_z=0.30$\,m, i.e., something closer than the expected static surface now occupies the pixel. In RGB-D mode the seeds are grown into a mask $M^{\mathrm{geo}}_t$ on the depth image (4-neighbours with depth difference $<\tau_g=0.02$\,m, at most 30\,px from the seed) and keypoints inside the mask are flagged. In stereo mode only keypoint depths are available, so keypoints within $r_g=15$\,px of a seed are flagged. Flagged keypoints are removed after the preliminary pose has been computed and before local-map tracking, so they enter neither the final pose optimisation nor new map points. \geom{} thus depends on depth availability, which in stereo mode is itself texture-dependent, and \flow{} depends on the conditioning of $\mathbf{F}_t$.

\textbf{Implementation and parameter choice.} Both filters were implemented for this study as a single additional source file in ORB-SLAM2's tracking thread (released with the patch); they are not the implementations used in the MSc dissertation. All parameters were fixed before the grid was run, using one texture-rich sequence (scene~1, L0/D1) only to check that the filters behaved as intended. On that sequence a first version (nearest-pixel $z'$, 60\,px growth limit) discarded 18\,\% of static keypoints in RGB-D mode (12\,\% with the edge-robust $z'$ alone); the 30\,px limit reduced this to 5\,\%. No parameter was changed after the grid was started.

\subsection{Controlled data generation}
\label{sec:data}
\textbf{Scenes and rendering.} We render $S=2$ furnished indoor scenes, an office ($7\times6\times2.8$\,m) and a laboratory/kitchen ($8\times5\times3$\,m), built from box and cylinder meshes with procedurally generated, tileable grey-scale textures (plaster, wood, tiles, brick, fabric, carpet, marble, posters, book spines). Images are produced by a purpose-written CPU ray caster (Open3D~0.20 \texttt{RaycastingScene}, Embree) with Lambertian shading from three point lights, ambient light, no shadows or global illumination, $2\times2$ supersampling, display gamma, and additive Gaussian noise ($\sigma=1$ grey level) whose realisation is identical in every condition. This renderer is far simpler than photo-realistic simulators; we chose it because it makes every factor exactly controllable and runs on a CPU. For each scene one camera trajectory (hand-held style, 15\,s at 30\,Hz, 450 frames, about 6.0\,m path length) is scripted and replayed identically for every texture and dynamics condition, so that viewpoint and scene geometry are held fixed.

\textbf{Texture levels.} Starting from the original materials (level L0), the albedo of every static surface is blended towards its per-material mean colour $\bar{A}$,
\begin{equation}
A_\ell(\mathbf{u})=(1-\beta_\ell)\,A(\mathbf{u})+\beta_\ell\,\bar{A},\quad \beta_\ell\in\{0,\beta_1,\beta_2,1\},
\end{equation}
for levels L0--L3 ($\beta_1=0.4$, $\beta_2=0.7$, chosen on 15 sub-sampled frames per scene so that $\NFAST$ falls to roughly one half and one fifth of L0); geometry, lighting and the dynamic agents' appearance are unchanged. Level L3 therefore retains only shading, geometric edges and object boundaries. Because $\beta_\ell$ is a rendering parameter rather than a perceptual quantity, each level is characterised by two image-based metrics computed on static pixels only (outside the ground-truth dynamic mask $M_t$): (i) the mean FAST corner count
\begin{equation}
\NFAST=\frac{1}{T}\sum_{t=1}^{T}\big|\{\mathbf{u}\notin M_t:\ \mathrm{FAST9}(I_t,\mathbf{u};t_F)\}\big|,
\end{equation}
with a fixed threshold $t_F=20$ on the full-resolution image and no adaptive lowering (unlike ORB-SLAM2's extractor, whose adaptive threshold would mask texture differences); and (ii) the mean gradient-magnitude entropy
\begin{equation}
\Hgrad=\frac{1}{T}\sum_{t=1}^{T}\Big(-\sum_{b=1}^{B}p_{t,b}\log_2 p_{t,b}\Big),
\end{equation}
where $p_{t,b}$ is the fraction of static pixels whose Sobel gradient magnitude $\|\nabla I_t(\mathbf{u})\|$ falls into bin $b$ of $B=64$ equal bins over $[0,g_{\max}]$, $g_{\max}=512$ (larger values fall into the last bin). Both are reported per level in Table~\ref{tab:texture}.

\textbf{Dynamic levels.} D0 has no moving agent, D1 one and D2 three. Agents are rigid, human-sized mannequins (1.75\,m, assembled from boxes, with strongly textured clothing) that walk at 0.9--1.2\,m/s along closed scripted paths; they are not articulated, and their textures are not changed with the texture level. The D1 agent is also present in D2, and every agent follows the same scripted path at every texture level. Because agent count does not fix image coverage, we report the mean dynamic-pixel ratio $\bar{\rho}=\frac{1}{T}\sum_t |M_t|/|\Omega|$ per sequence (Table~\ref{tab:texture}). The D0 cells serve as a control: any keypoint removed there is by construction a false positive.

\textbf{Sensors and ground truth.} Each sequence contains a rectified stereo pair (baseline $b_s=0.10$\,m, resolution $640\times480$, focal length 400\,px), a registered depth image of the left camera rendered from the z-buffer without a depth-noise model, ground-truth camera poses, and per-pixel dynamic masks. In total the design has $S\times4\times3=24$ sequences.

\begin{figure*}[t]
\centering
\includegraphics[width=\textwidth]{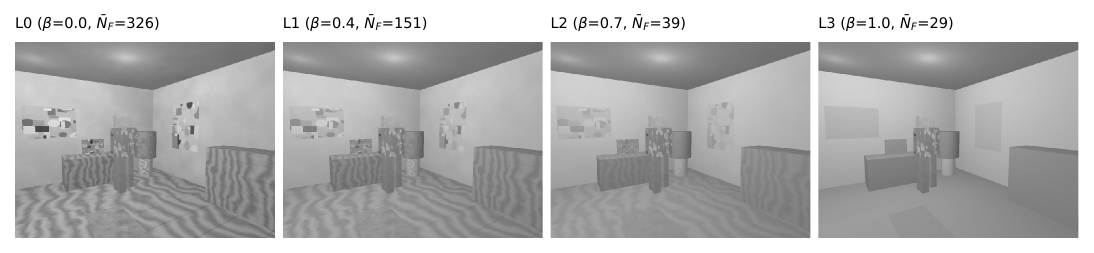}
\caption{Left-camera frames of scene~1 at texture levels L0--L3 (dynamic level D2, frame 180), rendered from the same pose. Titles give $\beta_\ell$ and the median $\NFAST$ over all sequences of that level.}
\label{fig:frames}
\end{figure*}

\begin{table}[t]
\caption{Texture and dynamics statistics per level (median over the six sequences of each level). Keypoints are ORB-SLAM2 left-image keypoints per frame in \none{} runs (identical for stereo and RGB-D input); the last column is the fraction of them with a valid stereo depth.}
\label{tab:texture}
\centering
\footnotesize
\begin{tabular}{@{}lccccc@{}}
\toprule
 & $\beta_\ell$ & $\NFAST$ & $\Hgrad$ [bits] & keypoints & stereo depth [\%] \\
\midrule
L0 & 0.0 & 326 & 2.65 & 1003 & 50 \\
L1 & 0.4 & 151 & 2.22 & 993 & 54 \\
L2 & 0.7 & 39 & 1.73 & 878 & 63 \\
L3 & 1.0 & 29 & 1.01 & 399 & 71 \\
\midrule
 & \multicolumn{5}{l}{Mean dynamic-pixel ratio $\bar{\rho}$:\quad D0: 0 \quad D1: 1.9\,\% \quad D2: 10.5\,\%} \\
\bottomrule
\end{tabular}
\end{table}

\subsection{Metrics}
\label{sec:metrics}
\textbf{Trajectory accuracy.} Let $\mathcal{T}$ be the set of frames for which the tracker output a pose, $\mathbf{P}_i$ the estimated and $\mathbf{Q}_i$ the ground-truth pose. Since stereo and RGB-D are metric, we align with the SE(3) transform $\mathbf{S}^\ast$ that minimises the translational residual~\cite{umeyama1991} (no scale), and compute
\begin{equation}
\mathrm{ATE}=\Big(\frac{1}{|\mathcal{T}|}\sum_{i\in\mathcal{T}}\big\|\mathrm{trans}(\mathbf{Q}_i^{-1}\mathbf{S}^\ast\mathbf{P}_i)\big\|^2\Big)^{1/2}.
\end{equation}
The translational RPE is computed over all pose pairs 1\,m apart along the trajectory. Both are computed with evo~\cite{grupp2017evo} (v1.37).

\textbf{Robustness.} ATE over tracked frames can hide failures~\cite{bonetto2023simulation}, so we also report the tracking completeness $C=|\mathcal{T}|/T$, the number of tracking losses and re-initialisations $N_{\mathrm{lost}}$, and the mean number of static keypoints used as tracking inliers per frame, $\bar{n}_s$. Runs with $C<C_{\min}=0.8$, and runs in which ORB-SLAM2 never initialised, are counted as failures; their ATE is excluded from the medians, and failures are reported next to every ATE value.

\textbf{Filter decisions.} A keypoint is ground-truth dynamic ($y_k=1$) if it lies inside $M_t$ (dilated by $\delta=2$\,px to absorb boundary effects). Over all frames of a run, with TP, FP, FN and TN counted on the keypoints of each frame before removal (frames in which a filter could not run, e.g.\ \geom{} before a map exists, count as all-static decisions),
\begin{equation}
\mathrm{P}=\tfrac{\mathrm{TP}}{\mathrm{TP}+\mathrm{FP}},\quad
\mathrm{R}=\tfrac{\mathrm{TP}}{\mathrm{TP}+\mathrm{FN}},\quad
\mathrm{FRR}=\tfrac{\mathrm{FP}}{\mathrm{FP}+\mathrm{TN}},
\end{equation}
where the static-feature false-removal rate FRR is the fraction of truly static keypoints that the filter discards. Such per-keypoint labels are not available in real benchmarks such as TUM RGB-D.

\subsection{Protocol and statistics}
\label{sec:protocol}
Because ORB-SLAM2 is not deterministic across runs, every configuration (scene $\times$ texture $\times$ dynamics $\times$ filter $\times$ sensor) is run $N_r=5$ times, giving 720 runs. We report medians over runs and scenes. The effect of filter $f$ in a cell is $\Delta\mathrm{ATE}_f=\mathrm{med}\,\mathrm{ATE}(f)-\mathrm{med}\,\mathrm{ATE}(\none)$; negative values mean the filter helps. Within a cell, run distributions are compared with the two-sided Wilcoxon rank-sum test~\cite{wilcoxon1945}; across cells, paired comparisons of cell medians use the Wilcoxon signed-rank test, and $p$-values are corrected with the Holm procedure~\cite{holm1979}. The trend of $\Delta\mathrm{ATE}_f$ with texture is summarised by Spearman's rank correlation with $\NFAST$ across per-scene cells (median over the five runs of one scene). All runs were executed on one Windows laptop (8 logical cores, 16\,GB RAM), three at a time; frames were fed in order without dropping, but tracking took 92\,ms per frame (median), i.e.\ slower than the 33\,ms frame period. Scripts, logs and raw trajectories are released (Sec.~\ref{sec:conclusion}).

\section{Experiments}
\label{sec:experiments}

\textbf{Hypotheses.} The design is built to test three pre-stated hypotheses; each may be refuted, and refutations will be reported as such.
\begin{itemize}
\item[\hyp{1}] The accuracy benefit of dynamic-point filtering ($-\Delta\mathrm{ATE}_f$) decreases as texture decreases, and may become negative at the lowest texture level, because filtering removes a larger share of the scarce static keypoints.
\item[\hyp{2}] \geom{} has a lower static-feature false-removal rate than \flow{}, and the gap widens as texture decreases, because $\mathbf{F}_t$ in Eq.~\eqref{eq:epi} is estimated from fewer and less well-distributed static correspondences; this difference in FRR, rather than a difference in recall, explains the accuracy gap between the filters.
\item[\hyp{3}] The interaction differs between sensors: in stereo mode, depth availability itself degrades with texture, so the advantage of \geom{} over \flow{} at low texture is smaller than in RGB-D mode, whose depth is texture-independent.
\end{itemize}

\subsection{Main grid: \texorpdfstring{texture $\times$ dynamics}{texture x dynamics}}
\label{sec:grid}
\begin{figure}[t]
\centering
\includegraphics[width=\columnwidth]{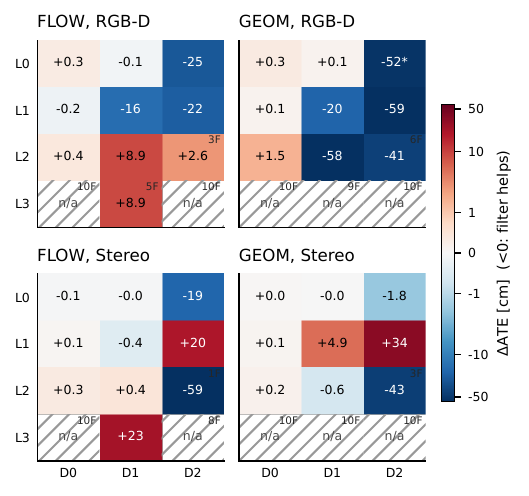}
\caption{Effect of filtering, $\Delta\mathrm{ATE}_f=\mathrm{med}\,\mathrm{ATE}(f)-\mathrm{med}\,\mathrm{ATE}(\none)$ in cm (medians over 2 scenes $\times$ 5 runs, failed runs excluded; negative = filter helps; symmetric-log colour scale centred at 0), per texture level (rows, L0 richest) and dynamic level (columns). $^\ast$: Holm-corrected Wilcoxon rank-sum $p<0.05$. $k$F: failed runs of the filtered configuration (out of 10). n/a: fewer than three successful runs of the filter or of \none{}.}
\label{fig:heatmap}
\end{figure}

Table~\ref{tab:main} lists median ATE, tracking completeness and failed runs for all cells, and Fig.~\ref{fig:heatmap} shows $\Delta\mathrm{ATE}_f$ as heat maps, one per filter and sensor, with texture on the vertical and dynamics on the horizontal axis. Completeness and failures are reported next to ATE so that an ATE improvement obtained by tracking fewer frames is not mistaken for a real gain. Under \hyp{1}, the heat maps would show the largest benefit at L0/D2 and a benefit shrinking towards L3; the D0 column isolates the pure cost of filtering, since there is nothing to remove.

Three observations frame the result. First, dynamics dominate: at L0 one agent costs \none{} only a few millimetres, but three agents (10.5\,\% of the pixels) raise its median ATE to 55\,cm in both modes, because the tracker partly follows the agents. Second, the pure cost of filtering is small: in the D0 column $|\Delta\mathrm{ATE}_f|\le1.5$\,cm and completeness is unchanged wherever ORB-SLAM2 initialises. Third, at L3 the static scene alone cannot carry ORB-SLAM2: L3/D0 frames yield only 233--288 ORB keypoints (median per scene), below the more than 500 that ORB-SLAM2's stereo/RGB-D initialisation requires, so none of the 60 L3/D0 runs initialised, and at L3/D1--D2 initialisation succeeded only because of the textured agents. Only one cell is significant after Holm correction (\geom{}, RGB-D, L0/D2: $-52$\,cm, $p_{\mathrm{Holm}}=0.017$); with ten runs per cell and bimodal outcomes (the tracker either ignores or follows an agent), the remaining differences are descriptive.

\textbf{\hyp{1} is not supported for accuracy and only partly supported for robustness.} Across the D1--D2 cells, the rank correlation between the benefit $-\Delta\mathrm{ATE}_f$ and $\NFAST$ was $\rho=0.49$ (\flow{}, RGB-D, $p=0.09$), $0.34$ (\flow{}, stereo, $p=0.23$), $-0.52$ (\geom{}, RGB-D, $p=0.10$) and $0.44$ (\geom{}, stereo, $p=0.18$); none is significant, and \geom{}'s RGB-D benefit was in fact largest at L1--L2 ($-58$\,cm at L2/D1). At L3, however, filtering lowered tracking completeness: relative to \none{}, the median $C$ in D1--D2 cells dropped by 0.39 (RGB-D) and 0.36 (stereo) with \geom{}, which produced a single successful L3 run, whereas \flow{} left $C$ essentially unchanged ($-0.01$) but did not improve ATE ($+9$\,cm RGB-D, $+23$\,cm stereo at L3/D1).

\begin{table*}[t]
\caption{Main grid. Each cell: median ATE [cm] over the successful runs of 2 scenes $\times$ 5 runs / median tracking completeness $C$ [\%] over all 10 runs / number of failed runs ($C<C_{\min}=0.8$ or no initialisation), for every texture level (rows) and dynamic level $\times$ filter (columns). --: fewer than three successful runs. RPE and per-scene values are given in the released result files.}
\label{tab:main}
\centering
\scriptsize
\setlength{\tabcolsep}{3pt}
\setlength{\tabcolsep}{2.2pt}
\begin{tabular}{@{}ll ccc ccc ccc@{}}
\toprule
 & & \multicolumn{3}{c}{D0 (static)} & \multicolumn{3}{c}{D1} & \multicolumn{3}{c}{D2} \\
\cmidrule(lr){3-5}\cmidrule(lr){6-8}\cmidrule(lr){9-11}
Sensor & Texture & \none & \flow & \geom & \none & \flow & \geom & \none & \flow & \geom \\
\midrule
\multirow{4}{*}{RGB-D} & L0 & 1.2/100/0 & 1.5/100/0 & 1.6/100/0 & 1.5/100/0 & 1.4/100/0 & 1.6/100/0 & 54.8/100/0 & 29.4/100/0 & 2.7/100/0 \\
  & L1 & 1.7/100/0 & 1.5/100/0 & 1.8/100/0 & 21.8/100/0 & 6.1/100/0 & 2.2/100/0 & 80.7/100/0 & 59.0/100/0 & 21.4/100/0 \\
  & L2 & 2.6/100/0 & 2.9/100/0 & 4.1/100/0 & 69.7/100/0 & 78.6/100/0 & 11.5/100/0 & 87.9/100/2 & 90.5/100/3 & 47.4/68/6 \\
  & L3 & --/0/10 & --/0/10 & --/0/10 & 57.5/73/5 & 66.4/73/5 & --/19/9 & --/62/9 & --/57/10 & --/9/10 \\
\midrule
 \multirow{4}{*}{Stereo} & L0 & 0.8/100/0 & 0.8/100/0 & 0.8/100/0 & 0.9/100/0 & 0.9/100/0 & 0.9/100/0 & 55.0/100/0 & 36.3/100/0 & 53.2/100/0 \\
  & L1 & 0.8/100/0 & 0.8/100/0 & 0.9/100/0 & 1.7/100/0 & 1.2/100/0 & 6.6/100/0 & 48.3/100/0 & 68.5/100/0 & 82.3/100/0 \\
  & L2 & 1.1/100/0 & 1.4/100/0 & 1.2/100/0 & 68.1/100/0 & 68.5/100/0 & 67.4/100/0 & 164/100/0 & 105/100/1 & 121/96/3 \\
  & L3 & --/0/10 & --/0/10 & --/0/10 & 94.7/100/0 & 118/99/0 & --/49/10 & 245/57/6 & --/57/8 & --/51/10 \\
\bottomrule
\end{tabular}
\end{table*}
\subsection{Mechanism: what the filters remove}
\label{sec:mechanism}
\begin{figure}[t]
\centering
\includegraphics[width=\columnwidth]{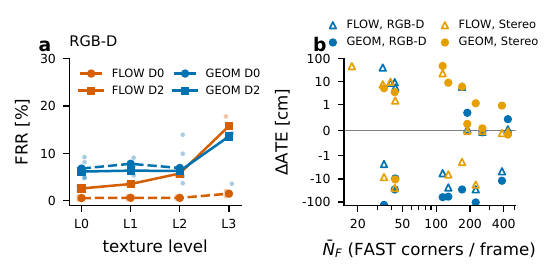}
\caption{(a) Static false-removal rate of \flow{} and \geom{} per texture level (D0 dashed, D2 solid; RGB-D; lines: median over 10 runs, dots: per-scene medians). \geom{} at L3/D0 is missing because ORB-SLAM2 never initialised there. (b) $\Delta\mathrm{ATE}_f$ against $\NFAST$ for stereo and RGB-D, one point per scene $\times$ texture $\times$ dynamics cell (D1--D2), symmetric-log axis; cells with fewer than three successful runs are omitted.}
\label{fig:mech}
\end{figure}

Table~\ref{tab:mechanism} reports, for the dynamic level D2 and for each texture level, the per-keypoint precision P, recall R and false-removal rate FRR of each filter, together with the surviving static inliers $\bar{n}_s$ (for \none{} this is the reference). The D0 cells add a pure-cost view: there, FRR is the fraction of all keypoints a filter removes for no reason. Fig.~\ref{fig:mech}(a) plots FRR against texture level. Under \hyp{2}, \flow{}'s FRR would rise more steeply than \geom{}'s as texture decreases, and $\Delta\mathrm{ATE}_f$ would be better predicted by FRR (or by $\bar{n}_s$) than by recall.

\textbf{\hyp{2} is not supported.} \geom{} discarded more static keypoints than \flow{}, not fewer: its FRR was higher in 86\,\% (RGB-D) and 95\,\% (stereo) of the scene $\times$ texture $\times$ dynamics cells in which both filters ran (median 6.8\,\% vs.\ 1.5\,\% for RGB-D and 19.6\,\% vs.\ 1.5\,\% for stereo). The predicted trend of \flow{} does appear: its FRR at D2 rose from 2.6\,\% (L0) to 15.8\,\% (L3) as $\mathbf{F}_t$ was estimated from fewer and less well-spread correspondences, so the FRR gap in favour of \flow{} narrowed and at L3 reversed (15.8\,\% vs.\ 13.6\,\% at D2; RGB-D: Spearman $\rho=-0.51$ between the gap and $\NFAST$, $p=0.014$). The accuracy differences, however, followed recall rather than FRR in RGB-D mode: across D1--D2 cells $\Delta\mathrm{ATE}_f$ correlated with R ($\rho=-0.49$, $p=0.016$) but not with FRR ($\rho=-0.11$, $p=0.60$); \geom{} found 56--59\,\% of the dynamic keypoints at L0--L1, \flow{} 15--20\,\%. In stereo mode the pattern reversed: $\Delta\mathrm{ATE}_f$ correlated with FRR ($\rho=0.40$, $p=0.046$) and not with R ($\rho=-0.12$), consistent with \geom{}'s high stereo FRR (caption of Table~\ref{tab:mechanism}).

\begin{table}[t]
\caption{Filter decisions at dynamic level D2: precision P, recall R and static false-removal rate FRR (\%), and static tracking inliers per frame $\bar{n}_s$; RGB-D mode, median over 2 scenes $\times$ 5 runs (runs in which the filter never ran are excluded). In stereo mode \geom{} has FRR 15--27\,\% and P 21--38\,\% at L0--L2 (released result files).}
\label{tab:mechanism}
\centering
\footnotesize
\setlength{\tabcolsep}{3.5pt}
\begin{tabular}{@{}l cccc cccc c@{}}
\toprule
 & \multicolumn{4}{c}{\flow} & \multicolumn{4}{c}{\geom} & \none \\
\cmidrule(lr){2-5}\cmidrule(lr){6-9}\cmidrule(lr){10-10}
 & P & R & FRR & $\bar{n}_s$ & P & R & FRR & $\bar{n}_s$ & $\bar{n}_s$ \\
\midrule
L0 & 66 & 20 & 2.6 & 271 & 71 & 56 & 6.2 & 275 & 270 \\
L1 & 61 & 15 & 3.5 & 234 & 77 & 59 & 6.4 & 252 & 224 \\
L2 & 56 & 10 & 5.8 & 133 & 75 & 28 & 6.3 & 153 & 132 \\
L3 & 49 & 7 & 15.8 & 61 & 71 & 20 & 13.6 & 96 & 59 \\
\bottomrule
\end{tabular}
\end{table}

\subsection{Stereo versus RGB-D}
\label{sec:sensor}
Stereo depth in ORB-SLAM2 comes from matching keypoints along epipolar lines and therefore requires texture, whereas rendered RGB-D depth does not. Fig.~\ref{fig:mech}(b) plots $\Delta\mathrm{ATE}_f$ against $\NFAST$ separately for both sensors, and Table~\ref{tab:texture} reports the fraction of keypoints with valid stereo depth per texture level. Under \hyp{3}, the advantage of \geom{} over \flow{} would decline with texture more strongly in stereo mode than in RGB-D mode.

\textbf{\hyp{3} is supported in outcome but not in mechanism.} In RGB-D mode the advantage of \geom{} over \flow{} ($\Delta\mathrm{ATE}_{\flow}-\Delta\mathrm{ATE}_{\geom}$, D1--D2 cells) grew as texture decreased (median 3.7, 35 and 38\,cm at L0, L1, L2; $\rho=-0.65$ with $\NFAST$, $p=0.03$), whereas in stereo mode it was absent or negative at every level ($-0.6$, $-12$, $-14$\,cm); at L3 neither filter had enough successful runs for a comparison. The hypothesised cause, however, is not what we observe: the fraction of keypoints with valid stereo depth did not fall with texture but rose from 50\,\% (L0) to 71\,\% (L3), because the keypoints that remain lie on high-contrast edges and on the agents. The stereo deficit of \geom{} is instead associated with its high static false-removal rate at all texture levels, which we conjecture stems from erroneous keyframe stereo depths entering the depth test (not tested here).

\section{Discussion and Limitations}
\label{sec:discussion}

\textbf{Relation to feature-scarcity-aware systems.} IL-SLAM, SR-SLAM and SFSE-SLAM~\cite{zhang2025ilslam,zhang2025srslam,li2026sfse} motivate their compensation mechanisms by the claim that dynamic-point removal leaves too few static features in low-texture regions. Our controlled data support a weaker form of this claim. At the lowest texture level filtering did reduce completeness, most strongly for \geom{}, with only about 60--100 static tracking inliers per frame left at L3/D2 (Table~\ref{tab:mechanism}). But the accuracy benefit of filtering did not shrink monotonically with texture, and at L2 it was often larger than at L0, because texture loss also made the unfiltered tracker more vulnerable to the agents (\none{} at L2/D1: 70\,cm vs.\ 1.5\,cm at L0, RGB-D). Moreover, the dominant low-texture failure in our grid occurs before any filter acts: ORB-SLAM2 did not initialise in static L3 scenes at all. A compensation mechanism that only adds features after removal would therefore not address the L3/D0 failures. Our results also contradict the common intuition, which we shared when stating \hyp{2}, that a geometric test is the conservative choice: \geom{} removed four to thirteen times more static keypoints than \flow{}, and it paid off only where its higher recall mattered and depth was dense and exact (RGB-D).

\textbf{Statistical strength.} Only one of 48 cells is significant after Holm correction, and none of the texture trends reaches $p<0.01$. The runs are bimodal (an agent is either ignored or followed), so cell medians over ten runs are unstable; the hypothesis tests above should be read as descriptive evidence on two scenes, not as general effect sizes.

\textbf{Synthetic-to-real gap.} Rendered images lack or simplify sensor noise, motion blur, rolling shutter, auto-exposure and depth artefacts, all of which interact with texture. Our renderer is deliberately simple (Lambertian, no shadows or global illumination, exact z-buffer depth), so effect sizes measured here may differ on real data, and the direction of effects is the primary quantity of interest.

\textbf{Single backbone and generic filters.} All results concern ORB-SLAM2 with semantics-free versions of two filtering principles. Systems with semantic masks, line or planar features, direct alignment or learned correspondences~\cite{bescos2018dynaslam,yu2018dsslam,yunus2021manhattan,teed2021droid,campos2021orbslam3} may respond differently to texture; our results do not transfer to them without further experiments.

\textbf{Few scenes.} With $S=2$ scenes and one trajectory each, differences between scenes cannot be separated from differences in, e.g., room size or furniture layout, and the main tables pool both scenes; per-scene cell values are released with the results.

\textbf{Texture metric.} $\NFAST$ depends on the FAST threshold and image resolution, and $\Hgrad$ ignores the spatial distribution of gradients; neither captures repetitive texture, which can cause false matches rather than a lack of matches. Albedo blending also leaves shading edges intact, so L3 is \emph{low}-texture rather than texture-free.

\textbf{Dynamics.} The dynamic levels are defined by agent count; image coverage $\bar{\rho}$ and agent speed also matter and are reported but not varied independently. The agents are rigid and keep their texture at every level, so at low texture they supply a growing share of all keypoints; this is realistic for textured clothing but confounds texture with the relative salience of the agents.

\textbf{Protocol.} Filter parameters were fixed on one texture-rich sequence of the grid (Sec.~\ref{sec:filters}); a held-out tuning scene would have been cleaner. Runs were executed three at a time and slower than real time; ORB-SLAM2's threads therefore had more time per frame than in a 30\,Hz deployment.

\section{Conclusion}
\label{sec:conclusion}

We have designed a controlled study that separates surface texture from scene dynamics along identical trajectories, in order to measure when dynamic-point filtering helps a feature-based SLAM front-end and why. In two scenes and 720 runs, dynamic-point filtering mattered far more than texture as long as ORB-SLAM2 could initialise, and its benefit did not decrease monotonically with texture (\hyp{1} not supported for accuracy); at the lowest texture level it reduced completeness instead. \geom{} was not the more conservative filter (\hyp{2} not supported): it discarded more static keypoints than \flow{}, and its RGB-D advantage came from higher recall. Its advantage disappeared in stereo mode, as \hyp{3} predicted, but not because stereo depth became sparse. The data-generation scripts, rendered sequences with ground-truth poses and masks, filter implementations and evaluation scripts are released so that other front-ends can be placed on the same grid.

\textbf{Data and code availability.} The renderer and scene definitions (all textures are procedural, no third-party assets), the ORB-SLAM2 patch with both filters, the run and evaluation scripts, per-run logs, trajectories and result tables are released at \url{https://github.com/felixxxue/texture-dynamics-slam}. Sequences are regenerated deterministically by the renderer. All code, including the ORB-SLAM2 patch (ORB-SLAM2 is GPLv3), is released under GPL-3.0; the results, logs and figures under CC BY 4.0.

\bibliographystyle{IEEEtran}
\bibliography{refs}

\end{document}